\pdfoutput=1

\documentclass[letterpaper, 10 pt, conference]{ieeeconf}  % Comment this line out if you need a4paper
\usepackage{times}
\usepackage{epsfig}
\usepackage{graphicx}
\usepackage{amsmath}
\newcommand{\norm}[1]{\left\lVert#1\right\rVert}
\usepackage{comment}
\usepackage{xcolor}
\usepackage{amssymb}
\usepackage{mathtools}
\usepackage{gensymb}
\usepackage{caption}
\usepackage{hyperref}

\IEEEoverridecommandlockouts                              % This command is only needed if 
\title{\LARGE \bf
    \centering{Spatial Semantic Embedding Network:} \\
	\centering{Fast 3D Instance Segmentation with Deep Metric Learning}
}

\author{Dongsu Zhang\\
	Seoul National University\\
	{\tt\small 96lives@snu.ac.kr}
	\and
	Junha Chun\\
	Seoul National University\\
	{\tt\small nikriz@snu.ac.kr}
	\and
	Sang Kyun Cha\\
	Seoul National University\\
	{\tt\small chask@snu.ac.kr}
	\and
	Young Min Kim\\
	Seoul National University\\
	{\tt\small youngmin.kim@snu.ac.kr}% <-this % stops a space
}

\begin{document}

\maketitle
\thispagestyle{empty}
\pagestyle{empty}

%%%%%%%%%%%%%%%%%%%%%%%%%%%%%%%%%%%%%%%%%%%%%%%%%%%%%%%%%%%%%%%%%%%%%%%%%%%%%%%%
\begin{abstract}
We propose spatial semantic embedding network (SSEN), a simple, yet efficient algorithm for 3D instance segmentation using deep metric learning.
The raw 3D reconstruction of an indoor environment suffers from occlusions, noise, and is produced without any meaningful distinction between individual entities.
For high-level intelligent tasks from a large scale scene, 3D instance segmentation recognizes individual instances of objects.
We approach the instance segmentation by simply learning the correct embedding space that maps individual instances of objects into distinct clusters that reflect both spatial and semantic information.
Unlike previous approaches that require complex pre-processing or post-processing, our implementation is compact and fast with competitive performance, maintaining scalability on large scenes with high resolution voxels.
We demonstrate the state-of-the-art performance of our algorithm in the ScanNet 3D instance segmentation benchmark on AP score.
\footnote{
	We plan on releasing the code for reproducibility.
	The code will be released on \url{https://github.com/96lives/ssen}
}

\end{abstract}

%%%%%%%%%%%%%%%%%%%%%%%%%%%%%%%%%%%%%%%%%%%%%%%%%%%%%%%%%%%%%%%%%%%%%%%%%%%%%%%%
\section{Introduction}

% why scene understanding is important
%The raw output of images or point cloud measurements is produced as a connected, and undistinguished piece of data. 
Scene understanding provides information on layout of multiple elements in the scene in terms of their locations, shapes, and poses in addition to possible relationship between them.
The extracted information is crucial to perform subsequent stages for intelligent applications, including high-level interaction or context-aware service~\cite{Zhoueaaw6661} in addition to autonomous driving, navigation, localization, or grasping. 
% Out  of  many  tasks  in  robotics,  the  scene  understanding provides  information  on  where  individual  objects  are. 
% Semantic or instance segmentation labels the elements of semantic entities, and from the information, the practical assistance of an intelligent agent is possible with high-level interaction or context-aware service~\cite{Zhoueaaw6661}.
Especially, 3D instance segmentation directly labels the measurement points in 3D as individual instances of objects, providing the exact location and extent of individual objects in the physical space.
% The result can serve as critical information for actual interaction on robots or natural user interface.

% overview of instance segmentation
3D instance segmentation started to gain attention relatively recently compared to the 2D instance segmentation, as the available deep learning architecture or datasets in 3D have been introduced in the past couple of years.
The first 3D instance segmentation approach was suggested by~\cite{wang2018sgpn} using the PointNet~\cite{qi2017pointnet} architecture on a rather small scale.
Larger-scale 3D scene understanding regresses bounding boxes of individual objects~\cite{bonet} or the centers of individual objects~\cite{qi2019votenet}.
For 3D instance segmentation, many recent works take advantage of 3D measurements for post-processing, mainly using the actual 3D shape of the proposed objects~\cite{yi2018gspn} or finding spatially connected components~\cite{Lahoud2019}. 
% While the three-dimensional  information  can  provide  important  information, it also requires additional post-processing steps.
On the other hand, our framework does not require additional post-processing steps demonstrating that learning the correct embedding is sufficient for instance segmentation.

\begin{figure}[t]
	\includegraphics[width=\columnwidth]{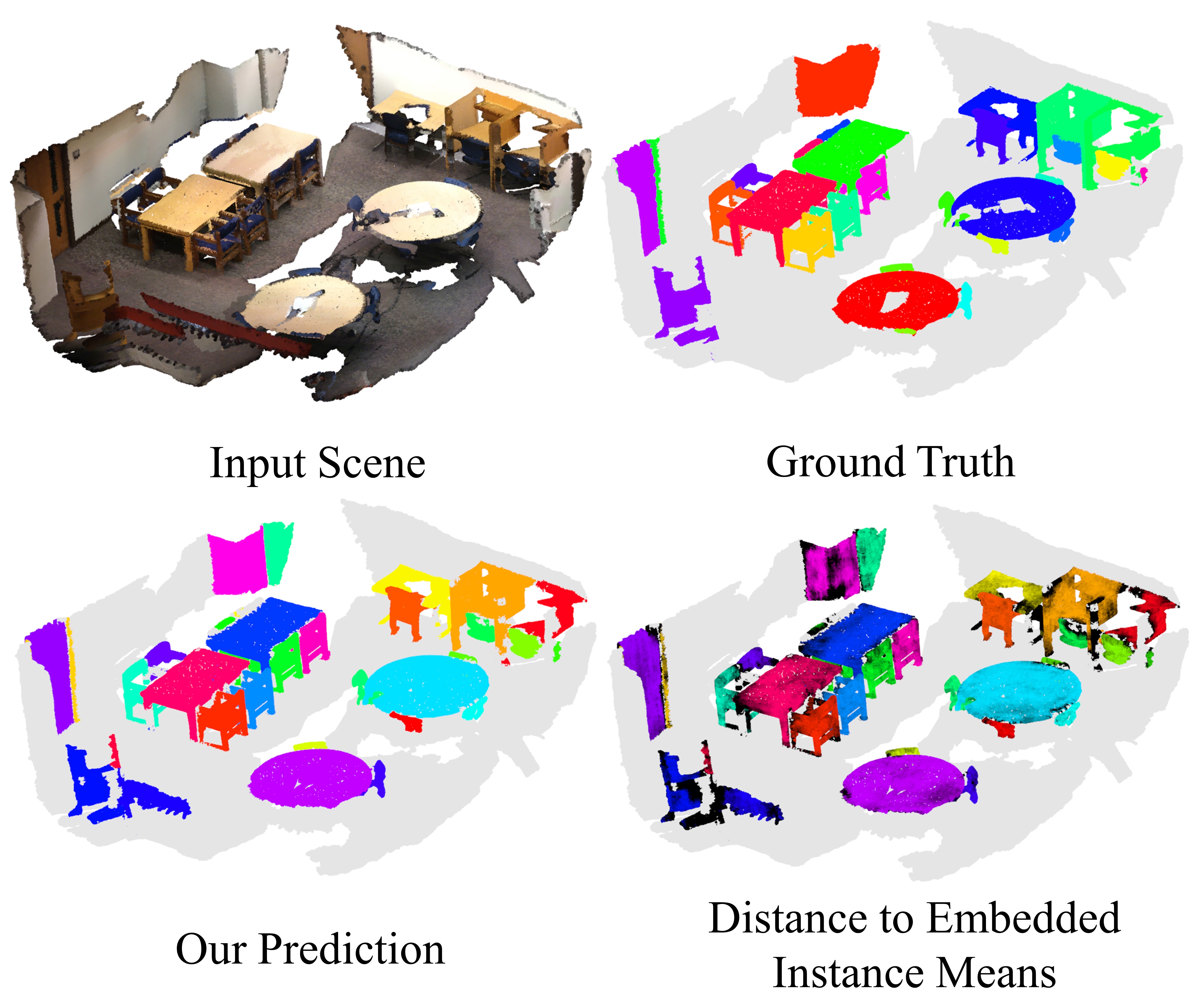}
	\caption{
		\textbf{Sample results of our method.} Given the input scene, our method outputs labels for individual object instances.
		%Different colors indicate different instances.
	    Our pipeline transforms the problem of instance segmentation into clustering in the embedded space.
	    The bottom right image visualizes the distance to the cluster mean in the embedded space. 
	    The black points represent embedding farther from the cluster mean, and we can observe that black points are physically located when nearby objects are in close contact with each other.
	    This shows us that our embedding captures not only semantic information of the scene but also spatial information.
	}
	\label{fig:mean_heatmap}
\end{figure}

% overview of our method
\par

We address the task of 3D semantic instance segmentation by training a deep neural network~\cite{minkowski,Graham2017} for metric learning, called \textbf{Spatial Semantic Embedding Network (SSEN)}.
The network is trained to embed each voxel to a feature space where embeddings from the same instance are closely located, while those from different instances are placed further away~\cite{DeBrabandere2017}.
As our algorithm effectively learns both semantic and spatial information, the instance segmentation of complex geometric shapes can be converted as a problem of simple clustering in the embedding space. 
We adapt to the different sizes of individual instances by applying the hierarchical density-based clustering~\cite{hdbscan} in the embedded space.
There is no post-processing in 3D Euclidean space, which is common in previous works, for example, iterations of multiple hypotheses, non-maximum suppression, or adaptive thresholds~\cite{Lahoud2019}.
As a consequence, our algorithm is more than two times faster than the state-of-the-art instance segmentation approach~\cite{bonet}. 

Our embedding space contains semantic information, robust to physical contact between objects, complex 3D topology or noise including occlusions while maintaining the overall spatial structure of the 3D scene, Fig.~\ref{fig:mean_heatmap}.
In short, our contribution can be summarized as the following:
(1) We suggest a highly scalable deep metric learning approach that turns an instance segmentation problem into a simple clustering in the embedding space which is the fastest method to our knowledge; and (2) we demonstrate that our algorithm, without any fine-tuning or explicit manipulation in the original 3D space, can achieve the state-of-the-art results in the ScanNet dataset on AP~\cite{dai2017scannet}.

\begin{figure*}[]
	\centering
	\includegraphics[width=0.9\textwidth]{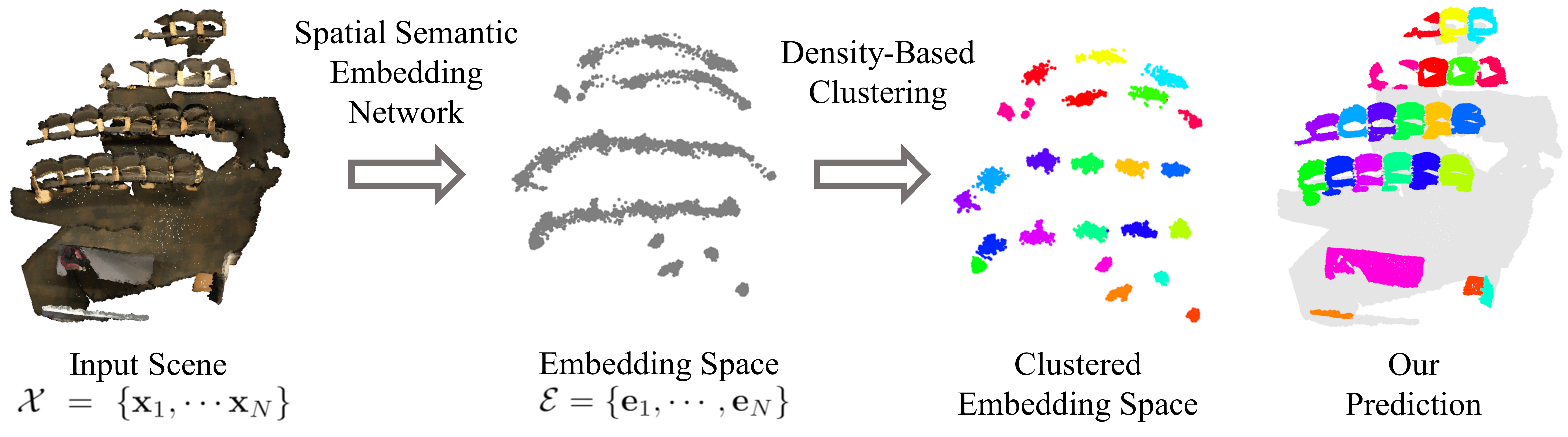}
	\caption{
		\textbf{Overall pipeline of our method.} Given the set of voxelized input $\mathcal{X}$,  the embedding network outputs the embedding of points $\mathcal{E}$. The embedded points are clustered using HDBSCAN algorithm, where the resulting clusters correspond to points that belong to individual instances.
	}
	\label{fig:model}
\end{figure*}

\section{Related Works}

We first review relevant previous works on 2D object recognition with a special focus on metric learning and instance segmentation.
For an extensive list of related works, we would like to refer to recent surveys on semantic segmentation~\cite{survey1,survey2} or metric learning~\cite{survey3,survey4}.
Then we describe neural network architecture for 3D data and the recent extensions into 3D instance segmentation.

\subsection{2D Object Recognition via Bounding Box Regression}
The mainstream method of object detection is to first create region proposals with bounding boxes and then to refine them while classifying the object inside the bounding box \cite{fast_rcnn, faster_rcnn, yolo, ssd}.
For instance segmentation, He et al.~\cite{he2017maskrcnn} propose a method to perform pixel-wise masking from obtaining the bounding box of the object.
While these methods result in the state-of-the-art performance for object recognition, nearby objects are sometimes misclassified  as a single instance of an object, due to the overlapping of bounding boxes. % TODO
These cases raise a question of whether methods based on the bounding box detection and refinement are suitable for object recognition~\cite{roi_prob}.

\subsection{2D Instance Segmentation via Metric Learning}
The approach by Fathi et al.~\cite{fathi} is one of the first papers to perform instance segmentation with deep metric learning.
They map each pixel to an embedding space where the pixels from the same instance are closely located while pixels from different instances are far from each other.
The neural network is trained extensively calculating the similarity for pairs of sampled pixels. 
A more scalable way to perform metric learning for instance segmentation is proposed by Branbandere et al.~\cite{DeBrabandere2017}.
They propose a discriminative loss function for deep metric learning that consists of inter-class loss and intra-class loss.
The inter-class loss pulls the embeddings of pixels of the same instances towards the means of the corresponding instance embeddings, or cluster centers.
The intra-class loss pushes the embedding centers of different instances far from each other.
This work is more efficient in terms of both memory and time compared to~\cite{fathi}, since no pairwise similarity between the pixels needs to be computed.
Also, this induces a far richer gradient using all the pixels in the image.

\subsection{Deep Neural Networks on Point Cloud}
While most of the deep learning approaches using 2D images are based on CNN-based architectures~\cite{Lecun98,alexnet2012}, 3D deep neural networks process different formats of 3D data and are limited in resolution.
Starting from PointNet~\cite{qi2017pointnet}, networks that directly process 3D point cloud measurements have been proposed.
The networks demonstrate good performance in segmentation or recognition in small-scale scenes with relatively simple structure~\cite{pointnetpp, dgcnn, pointcnn}.
However, recent works on sparse convolution \cite{sparseconv, minkowski} on voxelized point clouds have shown that neural networks in 3D data can be efficiently computed allowing the network to go deeper.
Not only is this method efficient in both memory and time, but sparse convolution~\cite{sparseconv, minkowski} has also outperformed all other point cloud neural networks on ScanNet semantic segmentation benchmark~\cite{dai2017scannet}.

\subsection{3D Point Cloud Instance Segmentation}
Compared to the recognition or segmentation for images, the instance segmentation in 3D data has acquired its attention lately with the recent advance of 3D deep learning architecture.
The work by Wang et al.~\cite{wang2018sgpn} is the first deep learning algorithm to perform instance segmentation on point cloud by grouping the points based on the similarity matrix.
% As 2D instance segmentation or object recognition utilizes bounding boxes or object regions, 3D instance segmentation approaches also consider the spatial context of individual objects including object proposals~\cite{yi2018gspn,3dsis} or bounding boxes~\cite{bonet}.
% On the other hand, we propose an approach that performs metric learning followed by clustering in the embedded space instead of post-processing in 3D space.
Wang et al. ~\cite{asis} and Pham et al. ~\cite{jsis} are the first to apply deep metric learning on 3D indoor instance segmentation.
However, both of them divided a large scene of point cloud into small overlapping windows to apply post-processings (such as k-means, CRF or mean-shift), which required a considerable time in merging the small windows.
Lahoud et al.~\cite{Lahoud2019} also try deep metric learning for 3D instance segmentation as a mean to obtain object proposals.
In addition to object proposals obtained by feature embedding, they also obtain object proposals using connected components and center predicting networks, followed by expensive non-maximum suppression.
Our work is closely related to the work by DeBrabandere et al. ~\cite{DeBrabandere2017,Lahoud2019}. 
However, we show that only obtaining embedding space is enough to perform instance segmentation outperforming all other methods on the ScanNet benchmark~\cite{dai2017scannet} on AP while being simple and fast.
%------------------------------------------------------------------------- 
\begin{table*}[ht]
\centering
\resizebox{\textwidth}{!}{
\begin{tabular} {c|ccccccccccccccccccc}
	$p$ value & \multicolumn{1}{c|}{AP} &
	bathtub & bed & bookshelf & cabinet & chair & counter & curtain & desk 
	& door & other & picture & refrig & showerCurt & sink & sofa & table & toilet & window \\
	\hline
	$p = 1$ & \multicolumn{1}{c|}{\textbf{0.302}} & \textbf{0.473} & 0.285 & \textbf{0.288} & 0.171 & 0.664 & 0.007 & \textbf{0.256} & \textbf{0.075}
	& 0.263 & 0.300 & 0.169 & 0.400 & 0.214 & 0.168 & \textbf{0.303} & 0.375 & \textbf{0.843} & 0.180 \\
	$p = 0.5$ & \multicolumn{1}{c|}{0.299} & 0.337 & \textbf{0.317} & 0.252 & 0.174 & 0.682 & 0.004 & 0.152 & 0.057
	& 0.272 & \textbf{0.304} & 0.174 & \textbf{0.426} & 0.258 & 0.279 & 0.255 & \textbf{0.399} & 0.826 & 0.209 \\
	$p = 0$ & \multicolumn{1}{c|}{0.292} & 0.303 & 0.299 & 0.223 & \textbf{0.191} & \textbf{0.694} & \textbf{0.010} & 0.131 & 0.055 
	& \textbf{0.298} & 0.279 & \textbf{0.188} & 0.368 & \textbf{0.292} & \textbf{0.281} & 0.258 & 0.342 & 0.812 & \textbf{0.223}
	\end{tabular}}
	\captionof{table}{
		\textbf{Average precision (AP) with different $p$ weighting for the loss function (Equation~(\ref{eq:L_inter}))} on validation set after training 50k steps.
		The objects with large scale such as bathtub, bookshelf, curtain, sofa have significantly higher AP when $p=1$.
		However, objects with a smaller number of points such as chair, sink, and windows tend to have higher AP when $p=0$.
	} 
	\label{tab:p_eff} 
\end{table*}

\section{Spatial Semantic Embedding Network}
%\subsection{Method Overview}
\label{sec:method}
Our system consists of two sparse convolution networks~\cite{sparseconv,minkowski}, one for semantic segmentation, and the other for obtaining the embedding of voxels, named \textit{Spatial Semantic Embedding Network} (SSEN).
The input to our system is voxels converted from a large-scale raw point cloud measurements of 3D scene.
The individual voxels have features $\mathbf{x}_k \in \mathbb{R}^f$ such as 3-D coordinates or colors.
We first use semantic segmentation network to obtain the labels of each voxels.
Since the backgrounds (floors and walls) are not of interest, we filter the points that are not predicted as backgrounds and input them to SSEN.
Note that this results in highly efficient training scheme, reducing the unnecessary points when training, maintaining high resolution voxels.

The overall pipeline of SSEN is depicted in Fig.~\ref{fig:model}.
SSEN converts the 3D measurements into embeddings, where individual instances are mapped far from each other by metric learning (Sec.~\ref{sec:loss}).
The SSEN maps the input of $N$ voxels $\mathcal{X}=\{\mathbf{x}_1, \cdots \mathbf{x}_N\}$ into $d$-dimension vectors $\mathcal{E}=\{\mathbf{e}_1,\cdots, \mathbf{e}_N\}$, where $\mathbf{e}_k \in \mathbb{R}^{d}$.
Then we run an HDBSCAN~\cite{hdbscan} on $\mathcal{E}$ where resulting clusters correspond to individual instances of objects (Sec.~\ref{sec:clustering}).
The semantic labels of resulting clusters are obtained from the semantic segmentation network.

\subsection{Loss Function for 3D Deep Metric Learning}\label{sec:loss}

Given the labels of individual instances of objects $i \in \mathcal{I}$, the set of $N$ embedded points $\mathcal{E}$ can be partitioned into individual instances, namely $\mathcal{E}_1, \cdots \mathcal{E}_{|\mathcal{I}|} \subset \mathcal{E}$, where $|\mathcal{I}|$ represents the number of instances existing in the given scene.
We train SSEN with the loss function combining the intra-instance loss $L_{intra}$ and the inter-instance loss $L_{inter}$ to learn the embedding.
% When we train  to learn the embedding for instance segmentation, we use the loss function combining the intra-instance loss $L_{intra}$ and the inter-instance loss $L_{inter}$
 %
\begin{equation}
L = L_{inter} + \gamma_{intra} L_{intra}
\label{eq:L_emb}
\end{equation}

The intra-instance loss $L_{intra}$ basically pushes the means $\mu_i$ of $\mathcal{E}_i$ in the embedded space far from each other.
\begin{equation}
L_{intra} = \frac{1}{|\mathcal{I}|(|\mathcal{I}| - 1)} 
\sum_{i \in \mathcal{I}} \sum_{\substack{j \in \mathcal{I} \\ j \neq i}}  [2\delta_{intra} - \norm{\mu_{i} - \mu_{j}} ]_+
\label{eq:L_intra}
\end{equation}
where $[x]_+ = \max(x, 0)$ denotes a hinge function.
The hinge function enables the repulsive force between clusters when the distance between the cluster centers is below a certain threshold, given as $2\delta_{intra}$.

On the other hand, the inter-instance loss pulls the points of the same instance together.
\begin{equation}
L_{inter} = \frac{1}{|\mathcal{I}|} 
\sum_{i \in \mathcal{I}} \frac{|\mathcal{E}_i|^p}{\sum_{j \in \mathcal{I}}|\mathcal{E}_j|^p} l_{i}
\label{eq:L_inter}
\end{equation}
where $l_{i}$ represents the average of inter-instance loss for the $i$-th instance.
\begin{equation}
l_{i} = \frac{1}{|\mathcal{E}_{i}|} 
\sum_{\mathbf{e}_{k} \in \mathcal{E}_i}[ \norm{\mu_i - \mathbf{e}_{k}} - \delta_{inter}]_+
\label{eq:L_i}
\end{equation}
Similar to the hinge function in Equation~(\ref{eq:L_intra}), the average loss of an individual instance $l_i$ is considered only for the elements that deviate farther than a threshold $\delta_{inter}$ from the cluster mean $\mu_i$.
Taking advantage of the mean instead of sampled pair-wise instances~\cite{DeBrabandere2017}, the calculation is efficient in terms of required computation and memory.
We did not use the regularization loss used in~\cite{DeBrabandere2017, Lahoud2019}, and empirically observed that the embeddings did not diverge.
The regularization loss limits the magnitude of instance means, but in spirit conflicts the intuition of metric learning that generates embedding with distant, isolated clusters.
% We did not use the regularization loss as in ~\cite{DeBrabandere2017, Lahoud2019} since it directly conflicted learning an embedding where means of instances were far from each other.
% We empirically found that the embeddings did not diverge without the regularization loss. 

While our loss function is based on the combination of attractive and repulsive forces as other metric learning approaches~\cite{DeBrabandere2017,Lahoud2019}, we generalize the formulation by adjusting weights for different sizes of instances.
Specifically, the average of inter-instance loss $l_{i}$ is weighted by $|\mathcal{E}_i|^p$ in Equation~(\ref{eq:L_inter}), where $|\mathcal{E}_i|$ represents the number of points that belong to the instance $i$ and $0 \leq p \leq 1$.
By controlling $p$, we can stabilize the training for objects of different 3D sizes, or equivalently, clusters with different numbers of points.

\par
If $p=0$, our formulation is equivalent to previous works~\cite{DeBrabandere2017, Lahoud2019}, where $l_i$ is summed up weighted equally.
This formulation is widely used in the image domain, where the number of pixels of an instance does not represent the actual scale of the physical instance, and therefore scale invariance is desired. 
In the 3D setting, however, the number voxels reflects the actual size of an object.
We observed that with $p=0$, the attractive force fails to condense embedded points that belong to the instances with larger sizes, while instances with a small number of points are well condensed.
On the other hand, if $p=1$, $L_{inter}$ is weighted equally for all points.
%In this case, the resulting embeddings of small clusters (hence small objects) are less effected.

%In short, increasing $p$ tends to segment big objects while decreasing $p$ tends to segment small objects well.
The argument is verified by comparing the performance of the instance segmentation  with $p=0$, $p=0.5$ and $p=1$ (Table~\ref{tab:p_eff}).
The AP is the highest at $p=1$ for large objects, such as bathtub, bookshelf, curtain, and sofa while $p=0$ performs the best for small instances or objects that often present occluded, for example, cabinet, chair, or counter.
For all of our presented results, we use $p=1$ which produces the best averaged performance for object instances of various scales.
%More results and discussions about the weighting of $p$ are presented in Section~\ref{subsec:p_eff}.

\subsection{Clustering in the Embedded Space}\label{sec:clustering}
After successful training, the same instance points lie close to each other separated by other instances.
The instance segmentation is mere clustering in the embedded space.
There are several criteria that need to be considered for the choice of the clustering algorithm:

\begin{itemize}
    \item 
    \textbf{
    	Clustering metric space induced by our loss function.
    }
    If our network is trained perfectly, all the embeddings of the same instances must be inside a sphere centered at $\mu_i$ with a radius of $\delta_{inter}$.
    The distances between means of different instances should be greater than or equal to $2\delta_{intra}$, and with triangular inequality, the distances between elements of different instances are at least $2\delta_{intra} - 2\delta_{inter} > 0$. 
    Intuitively, the resulted embedding space is dense around the mean within the pre-defined distance threshold, with varying density depending on the sizes of objects.
    %(We discuss further the density of the embeddings on the Ablation Study)
    \item
    \textbf{Clustering undetermined number of objects.}
    Since the number of instances is undetermined in the testing stage, the clustering algorithm should be able to predict the number of instances from the embedding space.
    This criterion makes $k$-means clustering~\cite{k_means} not applicable to our set-up.
    \item
    \textbf{Efficient and robust to outliers.}
    The clustering algorithm should be efficient and be able to robustly handle outliers in case the trained network fails to learn the proper embedding for a subset of points.
    Although the example in Fig.~\ref{fig:model} outputs embeddings that are dense near the mean of instance centers, outliers exist between nearby instances.
    The clustering algorithm should be able to handle that flexibly.
\end{itemize}{}

Out of various conventional clustering algorithms \cite{k_means, mean_shift, hdbscan, dbscan}, we select HDBSCAN~\cite{hdbscan} as our clustering algorithm. 
HDBSCAN is an extension of DBSCAN~\cite{dbscan} that uses hierarchical clustering.
HDBSCAN is very efficient and can robustly handle varying density clusters, and there is no need to fix the number of clusters.
On the other hand, mean-shift~\cite{mean_shift} was not scalable because multiple iterations are required, and DBSCAN performed poorly since it could not adapt to different densities.

%-------------------------------------------------------------------------

\begin{table*}[ht]
\centering
\resizebox{\textwidth}{!}{
\begin{tabular} {c|ccccccccccccccccccc}
	Method & \multicolumn{1}{c|}{AP} &
	bathtub & bed & bookshelf & cabinet & chair & counter & curtain & desk &
	door & other & picture & refrig & showerCurt & sink & sofa & table & toilet & window
	\\
	\hline
	
	\textbf{SSEN (with 10 rotations)} & \multicolumn{1}{c|}{\textbf{0.384}} & \textbf{0.852} & \textbf{0.494} & {0.192} & \textbf{0.226} & \textbf{0.648} & {0.022} & {0.398} & \textbf{0.299} & {0.277} & \textbf{0.317} & \textbf{0.231} & {0.194} & \textbf{0.514} & \textbf{0.196} & \textbf{0.586} & {0.444} & \textbf{0.843} & \textbf{0.184}\\
	
	\textbf{SSEN (without 10 rotation)} & \multicolumn{1}{c|}{{0.348}} & {0.778} & {0.432} & {0.214} & {0.213} & {0.635} & \textbf{0.038} & {0.280} & {0.275} & \textbf{0.288} & {0.297} & {0.185} & \textbf{0.259} & {0.324} & {0.188} & {0.425} & \textbf{0.450} & {0.812} & {0.178}\\
	
	\hline
	MTML &  \multicolumn{1}{c|}{0.282} & {0.577} & {0.380} & 0.182 & {0.107} & {0.430} & {0.001} &  \textbf{0.422} &  {0.057} &  {0.179} &  {0.162} &  {0.070} &  {0.229} &  {0.511} &  {0.161} &  {0.491} &  {0.313} &  {0.650} &  {0.162} \\
	
	MTML (FE) &  \multicolumn{1}{c|}{0.171}&  {0.235} &  {0.154} &  {0.075} &  {0.036} &  {0.475} &  {0.001} &  {0.114} &  {0.011} &  {0.142} &  {0.061} &  {0.028} &  {0.124} &  {0.392} &  {0.094} &  {0.172} &  {0.207} &  {0.636} &  {0.113} \\
	
	3D-BoNet &  \multicolumn{1}{c|}{0.253}&  {0.519} &  {0.324} &  {0.251} &  {0.137} &  {0.345} &  {0.031} &  {0.419} &  {0.069} &  {0.162} &  {0.131} &  {0.052} &  {0.202} &  {0.338} &  {0.147} &  {0.301} &  {0.303} &  {0.651} &  {0.178} \\
	
	PanopticFusion & \multicolumn{1}{c|}{0.214}&  {0.250} &  {0.330} &  \textbf{0.275} &  {0.103} &  {0.228} &  {0.000} &  {0.345} &  {0.024} &  {0.088} &  {0.203} &  {0.186} &  {0.167} &  {0.367} &  {0.125} &  {0.221} &  {0.112} &  {0.666} &  {0.162} \\
	
	MASC &  \multicolumn{1}{c|}{0.254}&  {0.463} &  {0.249} &  {0.113} &  {0.167} &  {0.412} &  {0.000} &  {0.374} &  {0.073} &  {0.173} &  {0.243} &  {0.130} &  {0.228} &  {0.368} &  {0.160} &  {0.356} &  {0.208} &  {0.711} &  {0.136} \\
	
	3D-SIS &  \multicolumn{1}{c|}{0.161}&  {0.407} &  {0.155} &  {0.068} &  {0.043} &  {0.346} &  {0.001} &  {0.134} &  {0.005} &  {0.088} &  {0.106} &  {0.037} &  {0.135} &  {0.321} &  {0.028} &  {0.339} &  {0.116} &  {0.466} &  {0.093}\\	
		
	DPC &  \multicolumn{1}{c|}{0.209}&  {0.403} &  {0.267} &  {0.230} &  {0.108} &  {0.283} &  {0.033} &  {0.278} &  {0.035} &  {0.088} &  {0.130} &  {0.034} &  {0.097} &  {0.229} &  {0.181} &  {0.273} &  {0.269} &  {0.695} &  {0.124}\\

\end{tabular}}
\captionof{table}{\textbf{AP.} The average precision (\textbf{AP}) score for ScanNet 3D instance segmentation benchmark. SSEN with 10 rotations indicate semantic labels obtained with 10 rotations and then averaged. Even without the 10 rotations, we still achieve far better results} \label{tab:AP} 

\end{table*} 
\begin{figure*}[ht]
\centering
	\includegraphics[width=\textwidth]{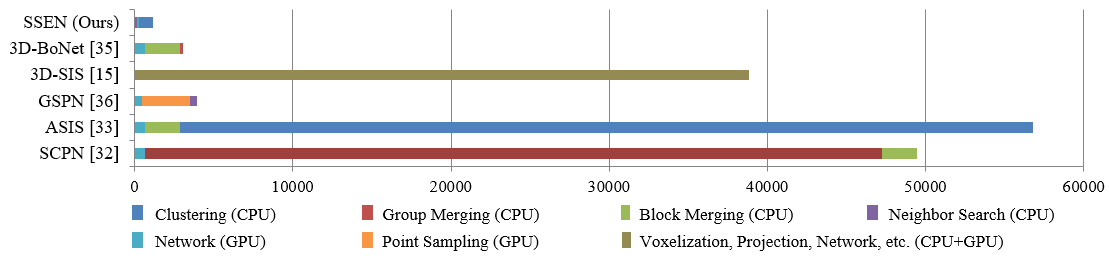}
	\caption{
	\textbf{Computational time} for various methods on the validation split (312 scenes) on ScanNet~\cite{dai2017scannet} (in seconds).
	The numbers except ours are excerpted from BoNet~\cite{bonet}.
	We are at least two times faster than BoNet~\cite{bonet} and about 50 times faster than ASIS~\cite{asis}.
	Our run time has been averaged from five runs, and the standard deviation is within 0.8\%.
	}
	\label{fig:comp_eff}
\end{figure*}

\section{Experimental Results}

\textbf{Implementation Details.}
The input scan is converted into voxels of size 2cm. 
The conversion uses the hash function implemented by Choy et al.~\cite{minkowski} with almost no overhead.
Individual voxels have the 3D coordinates and the RGB color value ($f=6$).
The output embedding dimension $d=8$, and we set the parameters of the loss function (Section~\ref{sec:loss}) as $\gamma_{intra}=10$, $\delta_{inter} = 0.1$, $\delta_{intra}=0.5$. 
We use the implementation of the Minkowski UNet34~\cite{minkowski} for both semantic segmentation network and SSEN.
% We used the ScanNet~\cite{dai2017scannet} to train, test and show various aspects of our algorithm.
% For data augmentation in the training phase, we randomly jitter the color of voxels with 0.03 scaled Gaussian noise.
% We also randomly scale the scene by $[0.8, 1.2]$.
% The scene is arbitrarily rotated around the $z$-axis, and around the $x$-axis with 5\degree scaled Gaussian noise, with maximum 10\degree.
We train our network using the ADAM optimizer~\cite{kingma:adam}, with the learning rate of 1e-4 for the batch size of 8.
The learning rate is decayed by 0.8 for every 10k iterations.
We use pretrained MinkowskiNet~\cite{minkowski} for semantic segmentation. 
% To filter background points for SSEN, we randomly use labels from ground truth or semantic segmentation network prediction.

\begin{figure*}[t]
\centering
	\includegraphics[width=0.9\textwidth]{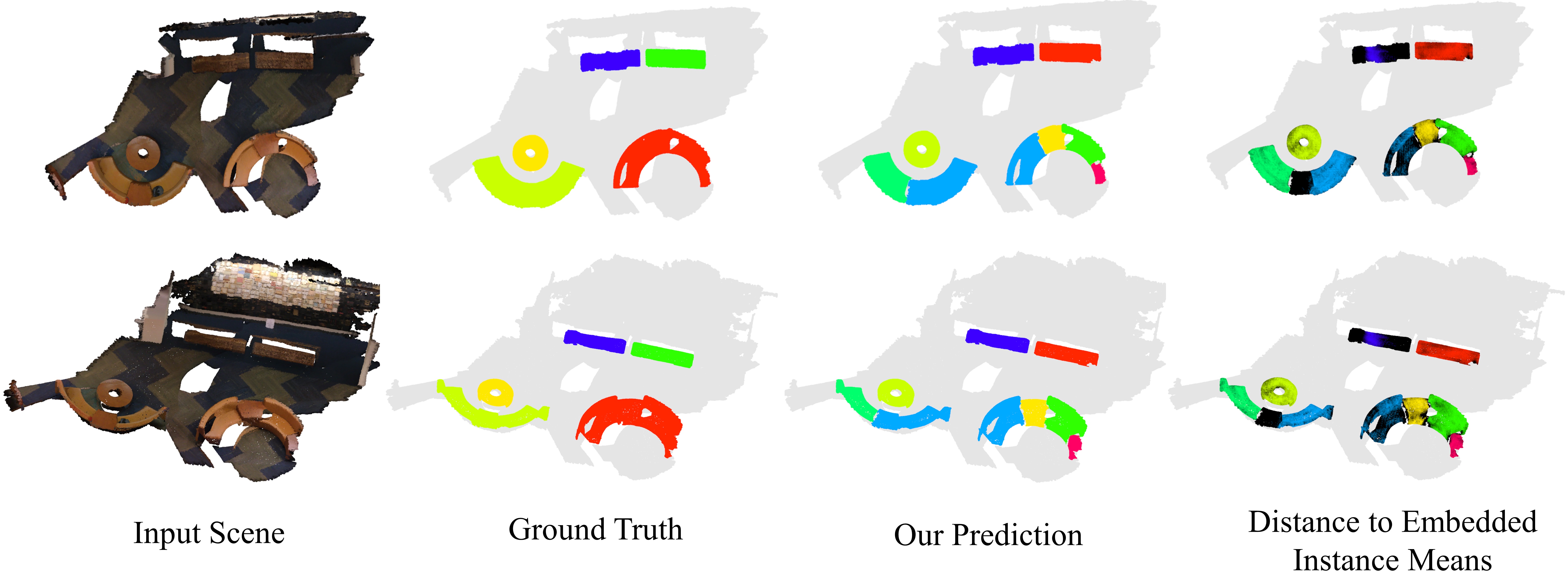}
	\caption{ \textbf{Sample failure cases of our method.} 
	%From left to right columns are input scene, ground truth instances, our instance predictions and the distance to embedded means.
	%If the points are black in the rightmost column, it specifies that the point is far from it's predicted instance's mean.
	The input scenes (first column) have two round couches that are composed of little couches.
	Although the ground truth is labeled them as connected (second column), our network predicts as if it were partitioned (third column).
	The last column visualizes the distance from the mean of the corresponding instance center with a color gradient from the cluster color (close to the cluster center) to black (far from the cluster center).
	}
	\label{fig:failure_cases}
\end{figure*}

\begin{figure*}[t]
	\centering
	\includegraphics[width=0.9\textwidth]{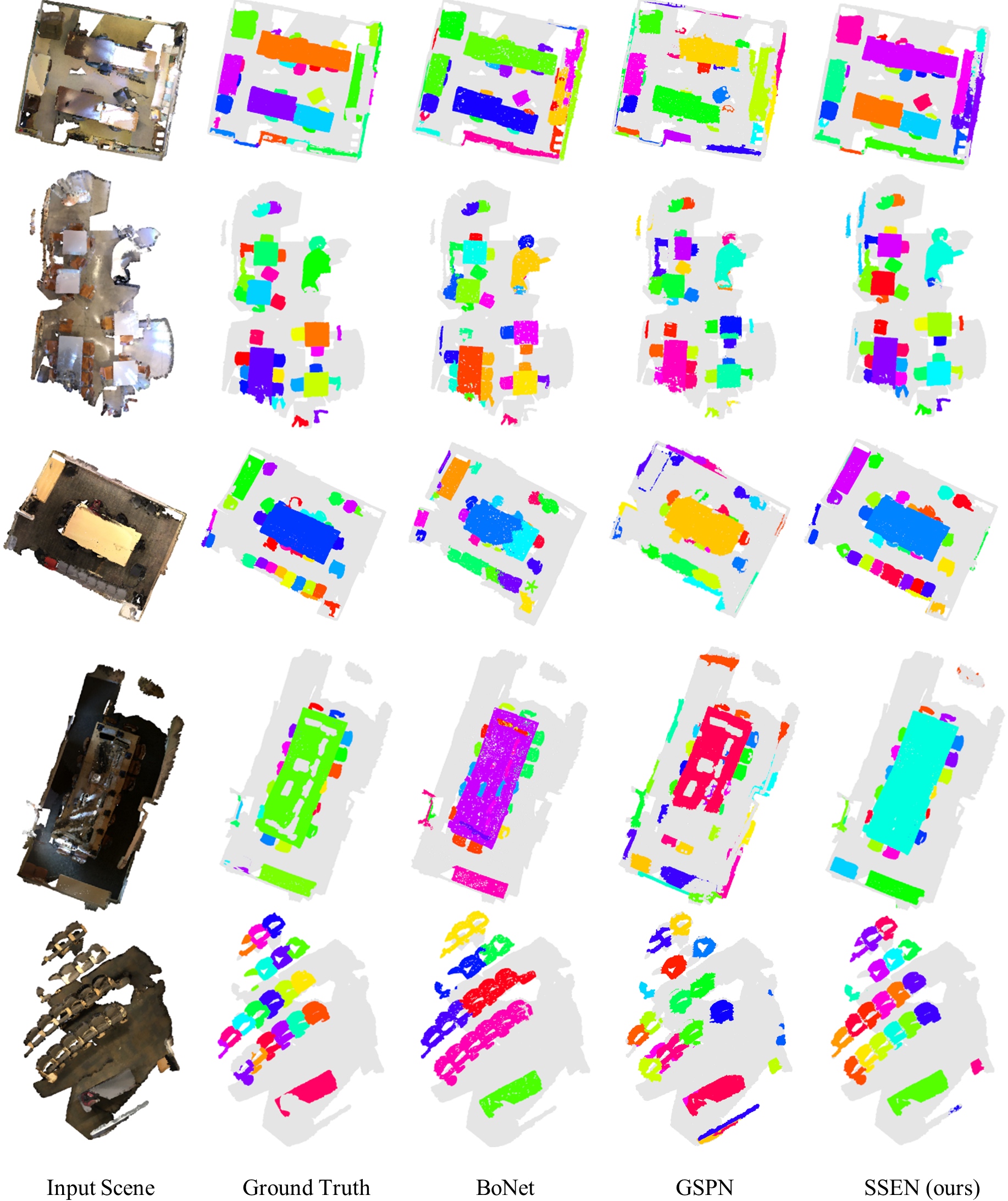}
	\caption{ \textbf{Qualitative comparison with other methods.} 
		We compare SSEN with other BoNet~\cite{bonet} and GSPN~\cite{yi2018gspn} on ScanNet~\cite{dai2017scannet} validation dataset.
		Our method is able to clearly distinguish different instances, predicting more precise instance labels than other methods. 
		(Different color indicates different instance, therefore the difference in color between ground truth and prediction is not important.)
	}
	\label{fig:appendix_qualitative}
\end{figure*}

% \textbf{Dataset.}
We used the ScanNet~\cite{dai2017scannet} to train, test and show various aspects of our algorithm.
For data augmentation in the training phase, we randomly jitter the color of voxels with 0.03 scaled Gaussian noise.
We also randomly scale the scene by $[0.8, 1.2]$.
The scene is arbitrarily rotated around the $z$-axis, and around the $x$-axis with 5\degree scaled Gaussian noise, with maximum 10\degree.

\begin{figure*}[ht!]
\centering
\includegraphics[width=0.9\textwidth]{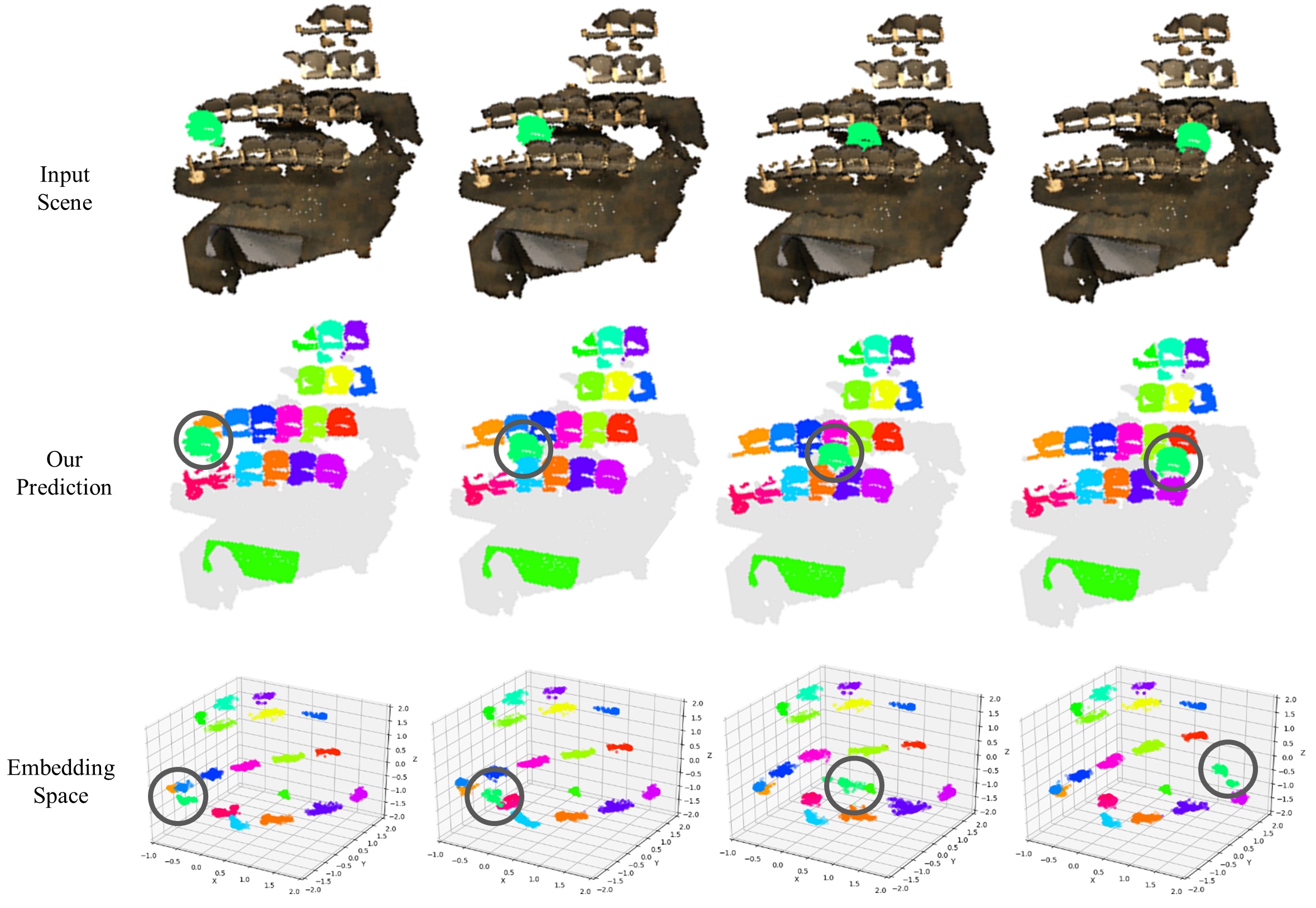}
\caption{\textbf{Spatial configuration and the embedding space.} 
We copied the set of point clouds that corresponds to a chair (circled, and highlighted in green) and moved it from left to right within a scene densely populated with multiple chairs. 
%The first, second and third row indicates the input scene, our prediction, and the embedding space, respectively.
%Given the manipulated input scene (top row), the configuration of corresponding clusters in the embedding space reflects the relative positions of individual instances.
%However, the new instance of chair pushes the existing neighboring instances introducing clear separation in the embedding spaces.
The continuous movement of clusters can be better observed in the accompanying video in the supplementary materials.
% The figures in each column are of the same scene.
% The first, second and the third row indicates the input scene, our prediction and the embedding space, respectively.
% 	The difference between each scenes is the location of a green point cloud chair instance circled.
% 	The physical movement of the corresponding instance is captured similarly within our metric space, correctly reflecting the relative positions of individual instances.
% 	Interestingly, the new instance of chair pushes away the existing instances in the embedding space while keeping the relative configuration, creating a clear cluster separated from neighboring instances.
% 	For the whole sequence of scenes with the chair moving, please refer to our video attached in the supplementary materials.
}
\label{fig:emb_vid}
\end{figure*}

\textbf{Evaluation on ScanNet Benchmark Challenge.}
ScanNet contains 3D reconstructed meshes of 1513 labeled indoor scenes for instance segmentation.
The benchmark challenge for 3D instance segmentation is then evaluated with unlabeled 100 test scans.
The performance is evaluated in terms of the average precision score.
% AP25 and AP50 are evaluation metrics denoting AP score with minimum intersection over union (IoU) score of 25\% and 50\% respectively.
The AP score averages scores obtained with IoU thresholds ranging from 50\% to 95\% with a step size of 5.

The quantitative results in Table~\ref{tab:AP} shows that we achieve the highest score for the average precision  compared to other state-of-the-art methods, namely: MTML~\cite{Lahoud2019}, 3D-BoNet~\cite{bonet}, PanopticFusion~\cite{panoptic}, MASC~\cite{masc}, 3D-SIS~\cite{3dsis}, and DPC~\cite{dpc}.
% The quantitative results are summarized in Table~\ref{tab:AP} with other state-of-the-art methods, namely: MTML~\cite{Lahoud2019}, 3D-BoNet~\cite{bonet}, PanopticFusion~\cite{panoptic}, MASC~\cite{masc}, 3D-SIS~\cite{3dsis}, and DPC~\cite{dpc}.
% We achieve the highest score for the average precision (Table~\ref{tab:AP}), and competitive results on AP50 (Table~\ref{tab:AP50}) and AP25 (Table~\ref{tab:AP25}).
This is remarkable considering that our algorithm does not use additional 3D information once the network maps the points into the embedding space, whereas other competitive results include post-processing stages.
In comparison, we outperform MTML (FE) by a large margin, which is the version of MTML~\cite{Lahoud2019} that uses only feature embedding without any post-processing using connected components.

We also carefully mention that the performance measure is partially degraded by the fact that our embedding can sometimes even learn more fine-grained instances than the instance labels are provided in the ground truth.
For example, in Fig.~\ref{fig:failure_cases} we show that our algorithm separately recognizes several instances of connected couches with little distinctive feature, which are grouped as a single instance in the ground truth.
While we can discuss this as a failure case, at the same, the result shows the discriminating power of our metric learning segmentation that can distinguish objects in complex configurations with interaction, occlusion or physical contact between objects.

As our formulation is compact and efficient, it has high potential to be employed in large-scale scenes for high-level tasks.
Figure~\ref{fig:comp_eff} compares the computational time of the entire pipeline for other compatible methods, excerpted from reported number on \cite{bonet}\footnote{We could not reproduce their results as the trained network was not available. The included numbers are reported in their original publication~\cite{bonet}, in which they ran on a Titan X GPU and the pre/post-processing steps on an i7 CPU core with a single thread. We run our algorithm on RTX 2080 GPU with i7 CPU core with a single thread. Note that our GPU is slightly faster, but we believe the comparison still is valid as our bottleneck is CPU.}. 
We are at least two times faster than that of BoNet which is the fastest algorithm reported to our knowledge and our method outperform all of the methods that are stated in performance.
As mentioned before, our algorithm does not add additional processing including group merging or block merging, which often consumes excessive power.
Also, our clustering algorithm (HDBSCAN~\cite{hdbscan}) is fast compared to the mean-shift~\cite{mean_shift} used for clustering in ASIS~\cite{asis}, which dominates the computation.
While our method only took 5 minutes to obtain results from ScanNet validation set (composed of 312 scans), replacing  HDBSCAN~\cite{hdbscan} with mean-shift~\cite{mean_shift} took one day with highly degraded performance.

Figure~\ref{fig:appendix_qualitative} presents the qualitative results of instance segmentation compared to BoNet~\cite{bonet} and GSPN~\cite{yi2018gspn}. 
Our superior performance is especially noticeable when the objects are in close proximity to each other.
For example, in the third and last row of the image, BoNet~\cite{bonet} and GSPN~\cite{yi2018gspn} fails to identify the aligned chairs as individual objects, marking them as a single instance or either not marking them all.
In comparison, our network clearly separates the nearby instances in the embedded space and distinguish them as different instances.

\textbf{Visualization of the Spatial-Semantic Embedding.}
We claim that our embedding space can capture semantic context while maintaining the spatial context of the input, and creates correct clusters for instance segmentation.
The effect of spatial context can be visualized by moving the 3D position of an instance, as shown in Fig.~\ref{fig:emb_vid}.
In this figure, we copy a single instance of the chair out of the populated chairs in the scene and move the chair along the empty row.
The chair is highlighted in green in the input scene (first row), and the corresponding segments in instance segmentation (middle row) and the embedding space with dimension 3 (last row) are circled.
Firstly, note that the chairs of the same rows are also aligned in a line in the embedding space with the same ordering.
 We can verify this by comparing the color patterns of prediction and the embedding space along the line.
 Also, every row of the 3D scene is mapped parallel to each other in the embedding space as the input scene.
 As the green chair moves between the two rows (image from the left column to right column), so does the embedding of the chair.
 This example gives us high confidence that our metric space resembles the spatial information of the input scene.
 Furthermore, as the chair moves, the whole embedding space changes pushing nearby instances away from the moving chair.
 This implies that the embedding captures the semantic information of the scene, where the different instances are further away while the points of the same instances are mapped together.
This demonstrates that our embedding represents not only the spatial information but the semantic information as well.

The mapping between the physical 3D space and the learned metric space can be also understood comparing the distance to the center of the clusters in embedding space.
Figure~\ref{fig:mean_heatmap} visualizes the distances to the cluster means for individual points, which also partially reflects the confidence of the clusters.
We can see that the 3D connectivity or other possible challenges can be effectively alleviated in the learned embedding space with densely clustered points.

%\subsection{Ablation Study}

%In this section, we study the effect of regularization parameter $p$.
%We argue that increasing $p$ tends to segment big objects while decreasing $p$ tends to segment small objects well.
%Table~\ref{tab:p_eff} compares the AP of network trained with $p=0$, $p=0.5$ and $p=1$. 
%The big object tends to have higher AP when $p=1$ and small objects tends to have small AP when $p=0$. 

%-------------------------------------------------------------------------
\section{Conclusion}
We present a simple, efficient metric learning approach for 3D instance segmentation.
Given a large-scale indoor scene, our deep network maps the voxels of the same instance object into a condensed cluster of embedded points.
The embedded space encodes both semantic information as well as the spatial structure of the original input.
The suggested pipeline outperforms all methods in both speed and performance. 
In the future, we would like to continue analyzing the embedded space.
Also, we will further explore real-time instance segmentation by removing the clustering stage, and training a network in an end-to-end fashion.

%Also, by removing the clustering and training the network from end-to-end fashion could lead us to real-time instance segmentation.

%%%%%%%-- supplementary : video, computational time, more qualitative results
%%%%%%%%%%%%%%%%%%%%%%%%%%%%%%%%%%%%%%%%%%%%%%%%%%%%%%%%%%%%%%%%%%%%%%%%%%%%%%%%

{
	\small
	\bibliographystyle{ieee_fullname}
	\bibliography{egbib}

\begin{thebibliography}{10}\itemsep=-1pt

\bibitem{hdbscan}
Ricardo J. G.~B. Campello, Davoud Moulavi, and Joerg Sander.
\newblock Density-based clustering based on hierarchical density estimates.
\newblock In Jian Pei, Vincent~S. Tseng, Longbing Cao, Hiroshi Motoda, and
  Guandong Xu, editors, {\em Advances in Knowledge Discovery and Data Mining},
  pages 160--172, Berlin, Heidelberg, 2013. Springer Berlin Heidelberg.

\bibitem{minkowski}
Christopher Choy, JunYoung Gwak, and Silvio Savarese.
\newblock 4d spatio-temporal convnets: Minkowski convolutional neural networks.
\newblock In {\em Proceedings of the IEEE Conference on Computer Vision and
  Pattern Recognition}, pages 3075--3084, 2019.

\bibitem{mean_shift}
Dorin Comaniciu and Peter Meer.
\newblock Mean shift: A robust approach toward feature space analysis.
\newblock {\em IEEE Trans. Pattern Anal. Mach. Intell.}, 24(5):603--619, May
  2002.

\bibitem{dai2017scannet}
Angela Dai, Angel~X. Chang, Manolis Savva, Maciej Halber, Thomas Funkhouser,
  and Matthias Nie{\ss}ner.
\newblock Scannet: Richly-annotated 3d reconstructions of indoor scenes.
\newblock In {\em Proc. Computer Vision and Pattern Recognition (CVPR), IEEE},
  2017.

\bibitem{DeBrabandere2017}
Bert {De Brabandere}, Davy Neven, and Luc {Van Gool}.
\newblock {Semantic Instance Segmentation with a Discriminative Loss Function}.
\newblock 2017.

\bibitem{dpc}
Francis Engelmann, Theodora Kontogianni, and Bastian Leibe.
\newblock Dilated point convolutions: On the receptive field of point
  convolutions.
\newblock {\em CoRR}, abs/1907.12046, 2019.

\bibitem{dbscan}
Martin Ester, Hans-Peter Kriegel, J\"{o}rg Sander, and Xiaowei Xu.
\newblock A density-based algorithm for discovering clusters a density-based
  algorithm for discovering clusters in large spatial databases with noise.
\newblock In {\em Proceedings of the Second International Conference on
  Knowledge Discovery and Data Mining}, KDD'96, pages 226--231. AAAI Press,
  1996.

\bibitem{fathi}
Alireza Fathi, Zbigniew Wojna, Vivek Rathod, Peng Wang, Hyun~Oh Song, Sergio
  Guadarrama, and Kevin~P Murphy.
\newblock {Semantic Instance Segmentation via Deep Metric Learning}.
\newblock Technical report.

\bibitem{survey2}
Alberto Garcia{-}Garcia, Sergio Orts{-}Escolano, Sergiu Oprea, Victor
  Villena{-}Martinez, and Jos{\'{e}}~Garc{\'{\i}}a Rodr{\'{\i}}guez.
\newblock A review on deep learning techniques applied to semantic
  segmentation.
\newblock {\em CoRR}, abs/1704.06857, 2017.

\bibitem{fast_rcnn}
Ross Girshick.
\newblock Fast r-cnn.
\newblock In {\em Proceedings of the 2015 IEEE International Conference on
  Computer Vision (ICCV)}, ICCV '15, pages 1440--1448, Washington, DC, USA,
  2015. IEEE Computer Society.

\bibitem{Graham2017}
Benjamin Graham, Martin Engelcke, and Laurens {Van Der Maaten}.
\newblock {3D Semantic Segmentation with Submanifold Sparse Convolutional
  Networks}.
\newblock Technical report, 2017.

\bibitem{sparseconv}
Benjamin Graham, Martin Engelcke, and Laurens van~der Maaten.
\newblock 3d semantic segmentation with submanifold sparse convolutional
  networks.
\newblock {\em CVPR}, 2018.

\bibitem{k_means}
J.~A. Hartigan and M.~A. Wong.
\newblock A k-means clustering algorithm.
\newblock {\em JSTOR: Applied Statistics}, 28(1):100--108, 1979.

\bibitem{he2017maskrcnn}
Kaiming He, Georgia Gkioxari, Piotr Doll\'{a}r, and Ross Girshick.
\newblock {Mask R-CNN}.
\newblock In {\em Proceedings of the International Conference on Computer
  Vision ({ICCV})}, 2017.

\bibitem{3dsis}
Hou Ji, Angela Dai, and Matthias Nie{\ss}ner.
\newblock 3d-sis: 3d semantic instance segmentation of rgb-d scans.
\newblock In {\em Proc. Computer Vision and Pattern Recognition (CVPR), IEEE},
  2019.

\bibitem{survey3}
Mahmut Kaya and H.s Bilge.
\newblock Deep metric learning: A survey.
\newblock {\em Symmetry}, 11:1066, 08 2019.

\bibitem{kingma:adam}
Diederick~P Kingma and Jimmy Ba.
\newblock Adam: A method for stochastic optimization.
\newblock In {\em International Conference on Learning Representations (ICLR)},
  2015.

\bibitem{alexnet2012}
Alex Krizhevsky, Ilya Sutskever, and Geoffrey~E Hinton.
\newblock Imagenet classification with deep convolutional neural networks.
\newblock In F. Pereira, C.~J.~C. Burges, L. Bottou, and K.~Q. Weinberger,
  editors, {\em Advances in Neural Information Processing Systems 25}, pages
  1097--1105. Curran Associates, Inc., 2012.

\bibitem{survey4}
Brian Kulis et~al.
\newblock Metric learning: A survey.
\newblock {\em Foundations and Trends{\textregistered} in Machine Learning},
  5(4):287--364, 2013.

\bibitem{Lahoud2019}
Jean Lahoud, Bernard Ghanem, Marc Pollefeys, and Martin~R Oswald.
\newblock {3D Instance Segmentation via Multi-task Metric Learning}.
\newblock 2019.

\bibitem{survey1}
Fahad Lateef and Yassine Ruichek.
\newblock Survey on semantic segmentation using deep learning techniques.
\newblock {\em Neurocomputing}, 338:321 -- 348, 2019.

\bibitem{Lecun98}
Yann Lecun, Léon Bottou, Yoshua Bengio, and Patrick Haffner.
\newblock Gradient-based learning applied to document recognition.
\newblock In {\em Proceedings of the IEEE}, pages 2278--2324, 1998.

\bibitem{pointcnn}
Yangyan Li, Rui Bu, Mingchao Sun, Wei Wu, Xinhan Di, and Baoquan Chen.
\newblock Pointcnn: Convolution on x-transformed points.
\newblock In S. Bengio, H. Wallach, H. Larochelle, K. Grauman, N. Cesa-Bianchi,
  and R. Garnett, editors, {\em Advances in Neural Information Processing
  Systems 31}, pages 820--830. Curran Associates, Inc., 2018.

\bibitem{masc}
Chen Liu and Yasutaka Furukawa.
\newblock {MASC:} multi-scale affinity with sparse convolution for 3d instance
  segmentation.
\newblock {\em CoRR}, abs/1902.04478, 2019.

\bibitem{ssd}
Wei Liu, Dragomir Anguelov, Dumitru Erhan, Christian Szegedy, Scott~E. Reed,
  Cheng-Yang Fu, and Alexander~C. Berg.
\newblock Ssd: Single shot multibox detector.
\newblock In {\em ECCV}, 2016.

\bibitem{panoptic}
Gaku Narita, Takashi Seno, Tomoya Ishikawa, and Yohsuke Kaji.
\newblock Panopticfusion: Online volumetric semantic mapping at the level of
  stuff and things, 03 2019.

\bibitem{jsis}
Quang-Hieu Pham, Duc~Thanh Nguyen, Binh-Son Hua, Gemma Roig, and Sai-Kit Yeung.
\newblock {JSIS3D}: Joint semantic-instance segmentation of 3d point clouds
  with multi-task pointwise networks and multi-value conditional random fields.
\newblock In {\em Proceedings of the IEEE Conference on Computer Vision and
  Pattern Recognition (CVPR)}, 2019.

\bibitem{qi2019votenet}
Charles~R Qi, Or Litany, Kaiming He, and Leonidas~J Guibas.
\newblock Deep hough voting for 3d object detection in point clouds.
\newblock In {\em Proceedings of the IEEE International Conference on Computer
  Vision}, 2019.

\bibitem{qi2017pointnet}
Charles~R. Qi, Hao Su, Kaichun Mo, and Leonidas~J. Guibas.
\newblock Pointnet: Deep learning on point sets for 3d classification and
  segmentation.
\newblock In {\em The IEEE Conference on Computer Vision and Pattern
  Recognition (CVPR)}, July 2017.

\bibitem{pointnetpp}
Charles~Ruizhongtai Qi, Li Yi, Hao Su, and Leonidas~J Guibas.
\newblock Pointnet++: Deep hierarchical feature learning on point sets in a
  metric space.
\newblock In I. Guyon, U.~V. Luxburg, S. Bengio, H. Wallach, R. Fergus, S.
  Vishwanathan, and R. Garnett, editors, {\em Advances in Neural Information
  Processing Systems 30}, pages 5099--5108. Curran Associates, Inc., 2017.

\bibitem{yolo}
Joseph Redmon, Santosh Divvala, Ross Girshick, and Ali Farhadi.
\newblock You only look once: Unified, real-time object detection, 2015.
\newblock cite arxiv:1506.02640.

\bibitem{faster_rcnn}
Shaoqing Ren, Kaiming He, Ross Girshick, and Jian Sun.
\newblock Faster r-cnn: Towards real-time object detection with region proposal
  networks.
\newblock In {\em Proceedings of the 28th International Conference on Neural
  Information Processing Systems - Volume 1}, NIPS'15, pages 91--99, Cambridge,
  MA, USA, 2015. MIT Press.

\bibitem{roi_prob}
C. {Shang}, Q. {Wu}, F. {Meng}, and L. {Xu}.
\newblock Instance segmentation by learning deep feature in embedding space.
\newblock In {\em 2019 IEEE International Conference on Image Processing
  (ICIP)}, pages 2444--2448, Sep. 2019.

\bibitem{wang2018sgpn}
Weiyue Wang, Ronald Yu, Qiangui Huang, and Ulrich Neumann.
\newblock Sgpn: Similarity group proposal network for 3d point cloud instance
  segmentation.
\newblock In {\em CVPR}, 2018.

\bibitem{asis}
Xinlong Wang, Shu Liu, Xiaoyong Shen, Chunhua Shen, and Jiaya Jia.
\newblock Associatively segmenting instances and semantics in point clouds.
\newblock In {\em The IEEE Conference on Computer Vision and Pattern
  Recognition (CVPR)}, June 2019.

\bibitem{dgcnn}
Yue Wang, Yongbin Sun, Ziwei Liu, Sanjay~E. Sarma, Michael~M. Bronstein, and
  Justin~M. Solomon.
\newblock Dynamic graph cnn for learning on point clouds.
\newblock {\em ACM Transactions on Graphics (TOG)}, 2019.

\bibitem{bonet}
Bo Yang, Jianan Wang, Ronald Clark, Qingyong Hu, Sen Wang, Andrew Markham, and
  Niki Trigoni.
\newblock Learning object bounding boxes for 3d instance segmentation on point
  clouds.
\newblock 2019.

\bibitem{yi2018gspn}
Li Yi, Wang Zhao, He Wang, Minhyuk Sung, and Leonidas Guibas.
\newblock Gspn: Generative shape proposal network for 3d instance segmentation
  in point cloud.
\newblock {\em arXiv preprint arXiv:1812.03320}, 2018.

\bibitem{Zhoueaaw6661}
Brady Zhou, Philipp Kr{\"a}henb{\"u}hl, and Vladlen Koltun.
\newblock Does computer vision matter for action?
\newblock {\em Science Robotics}, 4(30), 2019.

\end{thebibliography}
}

\end{document}